\def\year{2020}\relax
\documentclass[letterpaper]{article} 
\usepackage{aaai20}  
\usepackage{times}  
\usepackage{helvet} 
\usepackage{courier}  
\usepackage[hyphens]{url}  
\usepackage{graphicx} 
\usepackage{graphicx}  
\usepackage{amssymb}

\usepackage[]{todonotes}

\title{Detecting Logical Relation In Contract Clauses}
\author{\Large \textbf{Alexandre Yukio Ichida}\\ \Large \textbf{Felipe Meneguzzi} \\ 
Pontifical Catholic University of Rio Grande do Sul\\ 
 Porto Alegre, RS, Brazil\\
 alexandre.ichida@edu.pucrs.br\\
 felipe.meneguzzi@pucrs.br
 }
 
\begin{document}

\maketitle

\begin{abstract}
Contracts underlie most modern commercial transactions defining define the duties and obligations of the related parties in an agreement. 
Ensuring such agreements are error free is crucial for modern society and their analysis of a contract requires understanding the logical relations between clauses and identifying potential contradictions. 
This analysis depends on error-prone human effort to understand each contract clause. 
In this work, we develop an approach to automate the extraction of logical relations between clauses in a contract.
We address this problem as a Natural Language Inference task to detect the entailment type between two clauses in a contract. 
The resulting approach should help contract authors detecting potential logical conflicts between clauses. 
\end{abstract}


\section{\label{chap:intro}Introduction}

Understanding existing logical relations between sentences is a difficult task that requires an accurate understanding of natural language meaning.
The ambiguity and variability of linguistic expression in natural language complicates the recognition of these relations such as entailment and contradiction contained in texts.
The ability to classify these logical inferences among different text is a significant feature of an intelligent system~\cite{bos2005recognising}. 
Detecting these logical relations can help humans to interpret a more complex text, where entailment and contradiction are crucial aspects to fully understanding such as norms and contracts.
%
Contracts are documents that contain normative sentences formalizing agreements among the related parties, which involve people and companies.
The normative sentences describe the duties that the related parties are subject to and the penalties in case of rule violation.
In a contract, the norms may be logically related, so that some are entailed, or contradict each other~\cite{AIRESDM15}.

For instance, in a contract that contains the following norms ``All companies must pay the Y tax'' and ``The company X must pay the Y tax'', it is not possible the first norm is satisfied while the second norm is not satisfied.
In the case of company X not paying the tax Y, automatically violates both norms due to the conditions of compliance.
Considering that both norms are logically linked and are in the same context, we have an entailment relation between them.
By contrast, conflicts in a contract may emerge through problems related to a logical contradiction between norm clauses.
Taking the example above, we have a contradiction relation if we change the second norm to ``The company X must not pay the tax Y'' due to their contradictory compliance condition.
Analyzing these conflicts in a contract demands a careful analysis from both related parties.
An automated way to detect a conflict between contract clauses addresses these reviews of contract clauses, which is a long and complex issue even for human experts.

Classifying the logical relation between norms is analogous to Natural Language Inference (NLI), which is the task of determining whether a natural language hypothesis \textit{h} can be inferred from a natural language premise \textit{p}~\cite{Maccartney:2009:NLI:1751277}.
In an entailment relation, if \textit{p} is true then \textit{h} cannot be false, otherwise there is a contradiction. 
NLI is a broader task than conflict identification, and thus, good models to classify logical relations will naturally be applicable to detect contract conflicts. 
Importantly, since NLI has seen a surge in research, including new machine learning models and dataset curation~\cite{bowman:2015:snli,mnli-N18-1101}, it offers substantial labelled training data in much larger quantities than contract conflict datasets~\cite{aires_joao_paulo_2017_345411}. 

We develop an automated approach to detecting logical relations between norms in a contract as a natural language inference problem.
First, we develop a neural network that addresses the Natural Language Inference task to classify the logical relation between two sentences written in natural language.
Second, we apply the trained neural network on conflicting norm pairs reporting the logical relation that our model predicts concerning the conflict types. 
The resulting model can help identify potential inconsistencies between contract clauses by detecting logical relationships between natural language sentences.






\section{Natural Language Inference}
Automated reasoning and logical inference studied in natural logic are important topics of artificial intelligence. 
Natural language inference (NLI) is a widely-studied natural language processing task that is concerned with determining the inferential relation between a premise \textit{p} and a hypothesis \textit{h} \cite{bowman:2015:snli}.
In NLI, the entailment relation inferred is formulated based on the following representations: two-way classification and three-way classification \cite{Maccartney:2009:NLI:1751277}. 

Two-way classification is the simplest representation of NLI, which describes the task as a binary decision.
The objective of this NLI task is to classify whether the hypothesis follows the premise (entailment) or does not (non-entailment).
%
Alternatively, 
in three-way classification form, the relations are divided into three categories: entailment, contradiction and neutral.
Given a pair of premise-hypothesis \textit{p} and \textit{h}, the \textit{entailment} relation occurs when \textit{h} can be inferred from \textit{p}~\cite{bowman:2015:snli}. 
When \textit{h} entails the negation of \textit{p}, the pair results in a \textit{contradiction}.
Otherwise, if none of these relations can be inferred, the relation of \textit{p} and \textit{h} is \textit{neutral}.

In NLI, both \textit{p} and \textit{h} are sentences written in natural language.
The challenge of this task differs of formal deduction from logic due to its focus in informal reasoning~\cite{Maccartney:2009:NLI:1751277}.
The emphasis of the NLI is on aspects of natural language such as lexical semantic knowledge and the deal with the variability of linguistic expression.
Consider the following premise \textit{p} and hypothesis \textit{h} as an instance of an NLI scenario~\cite{Maccartney:2009:NLI:1751277}:

\begin{itemize}
    \item \textit{p}: Several airlines polled saw costs grow more than expected, even after adjusting for inflation.
    \item \textit{h}: Some of the companies in the poll reported cost increases.
\end{itemize}

This example is considered a valid entailment inference in the NLI context because any person that interprets \textit{p} would likely accept that \textit{h} implies the information of \textit{p}.
Although this is a valid NLI classification, \textit{h} is not a strict logical consequence of \textit{p} due to the fact that \textit{p} informs that airline companies \textbf{saw} the growth of the cost, not necessarily \textbf{reporting} the growth of the cost. 
This example reflects the informal reasoning of the task definition due to deal with ambiguity of natural language~\cite{Maccartney:2009:NLI:1751277}.

\section{Natural Language Inference Classifier}
In this section, we explain our NLI model to predict over normative sentences.
First, we explain the Transformer neural network architecture that we use to deal with NLI task.
Second, we detail the attention mechanism we use to help our classifier focus on key parts of normative sentences.
Third, we describe the feed-forward neural network that refines the internal representation of words in sentences.
Finally, we show how the model predicts a class given its output representation.

\subsection{Transformer}
Transformer is a type of neural network architecture that processes sequences based solely on attention mechanisms instead of using recurrent connections in the network.
Vaswani~\textit{et al.}~\cite{NIPS2017_7181} developed this architecture to deal with the machine translation task, achieving the state-of-the-art performance.
This architecture uses an Encoder-Decoder approach based on other machine translation neural networks such as Sequence to Sequence learning~\cite{sutskever2014sequence}.
Approaches that use Transformer variations recently achieve state-of-the-art results on other natural language processing tasks, such as Question Answering~\cite{devlin2018bert} and Sentiment Classification~\cite{devlin2018bert}.

\begin{figure}
\centering
  \includegraphics[width=0.4\textwidth]{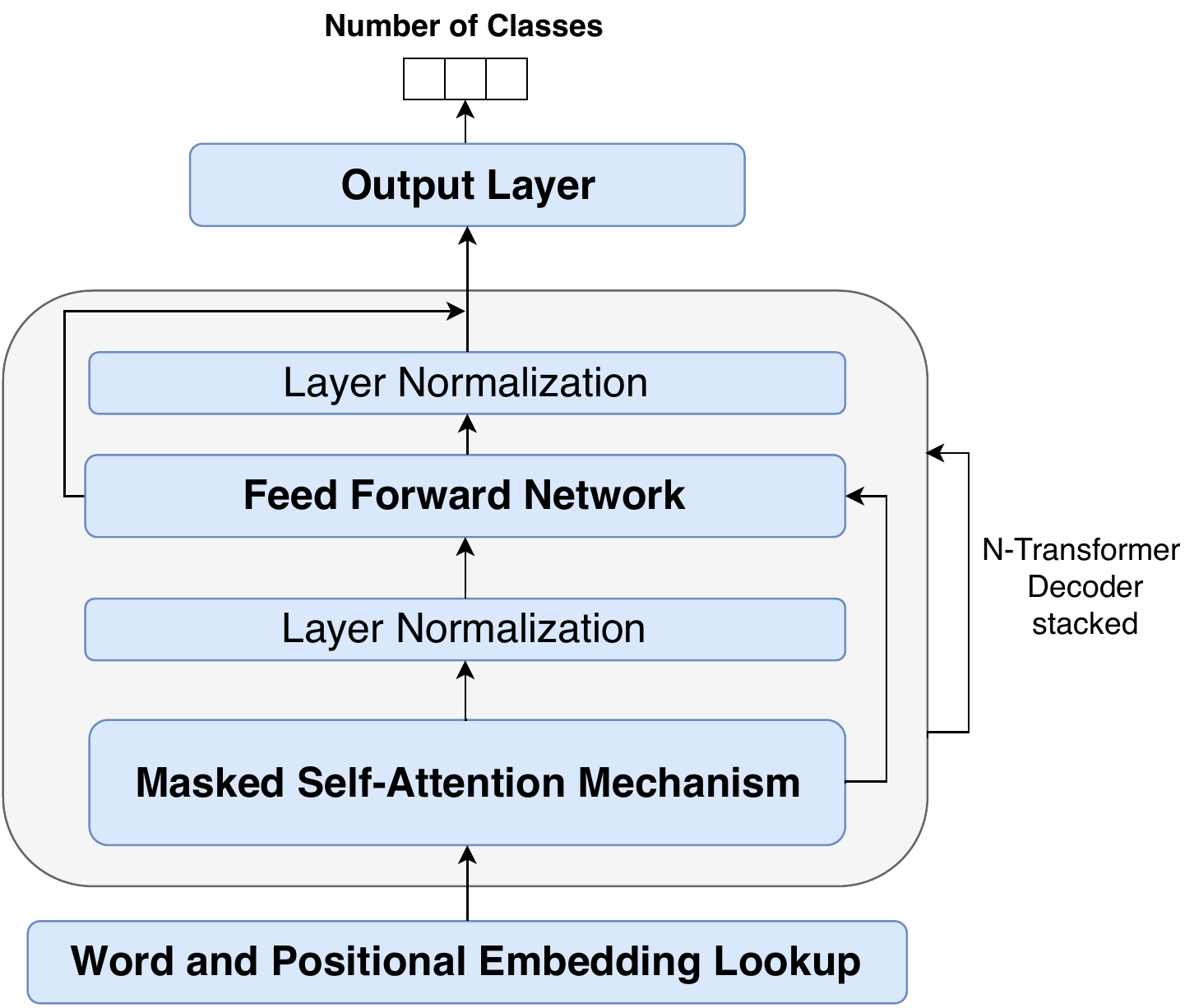}
\caption{Diagram of a Transformer Decoder block that may contain more than one stacked block in a single model.}
\label{fig:decoder-block}       
\end{figure}

Instead of using the entire Transformer architecture from machine translation tasks, we use only the Decoder part based on the work by Radford~\textit{et al.}~\cite{radford2018improving} that deals with the NLI classification task.
The Transformer Decoder contains blocks that consist of a layer with a Self Attention Mechanism and a feed-forward neural network module.
In the decoder block, we add a residual connection applying a sum over the input of each layer with its output followed by layer normalization.
Figure~\ref{fig:decoder-block} illustrates this overall architecture with details of Decoder Transformer block of our model and how we adapt to a classification task.

\subsubsection{Self-Attention}
Self-attention is a neural network mechanism that relates different positions of a single sequence to create a representation of the sequence.
In natural language processing models, the objective of self-attention mechanism is to focus on relevant words rather than all words of the input sentence, giving an attention score for each word relation of the sequence.
For example, the word "not" is more relevant to check whether exists some contradiction between two sentences.
The self-attention mechanism in a Transformer calculates the attention of each word using a query that maps an output given a set of key-value pairs~\cite{NIPS2017_7181}.
Initially, our model uses a fully connected layer on the input representation and then split its output into three matrices, which represent the initial \textit{query}, \textit{key} and \textit{value} matrices of our attention model.

Our neural network uses the Scale Dot-Product attention model, which applies a matrix multiplication between the \textit{query} and \textit{key} matrices followed by a softmax function with the \textit{value} matrix.
To obtain more stable gradients, the attention model scales the matrix multiplication dividing it by $\sqrt{d_k}$, where $d_k$ represents the dimension of \textit{key} matrix.
Equation~\ref{eq:scale-dot-product} computes the Scale Dot-Product attention model given a \textit{query}, \textit{key}, and \textit{value} matrices represented by \textit{Q}, \textit{K} and \textit{V} respectively.
\begin{equation} \label{eq:scale-dot-product}
    softmax(mask(\frac{Q.K}{\sqrt{d_K}})).V
\end{equation}

To prevent the self-attention mechanism from computing the attention score to subsequent positions of a single word, we apply a mask that removes the score of subsequent words.
Given a certain word, we put a $-\infty$ value in its subsequent words in input of softmax generating extremely low attention scores in these illegal positions~\cite{NIPS2017_7181}.
This masking process ensures that a word can depend only on already seen ones \cite{NIPS2017_7181} ignoring its subsequent words. 
Table~\ref{tab:attention-mask} describes an example of attention masking while processing the word ``being'' in sentence pair. 
\begin{table*}[!htb]
\centering
\caption{Example of sentence pair masked that contains very low values ($-\infty$) on subsequent positions.}
\label{tab:attention-mask}
\begin{tabular}{lllllllllll}
A & Car & is & \textbf{being} & \textit{driven} & \textit{.} & \textit{A} & \textit{Car} & \textit{is} & \textit{stuck} & \textit{[EOS]} \\ \hline
\multicolumn{1}{|c|}{0.5} & \multicolumn{1}{c|}{0.78} & \multicolumn{1}{c|}{0.1} & \multicolumn{1}{c|}{current} & \multicolumn{1}{c|}{$-\infty$} & \multicolumn{1}{c|}{$-\infty$} & \multicolumn{1}{c|}{$-\infty$} & \multicolumn{1}{c|}{$-\infty$} & \multicolumn{1}{c|}{$-\infty$} & \multicolumn{1}{c|}{$-\infty$} & \multicolumn{1}{c|}{$-\infty$} \\ \hline
\end{tabular}
\end{table*}

\begin{figure}
\centering
  \includegraphics[width=0.4\textwidth]{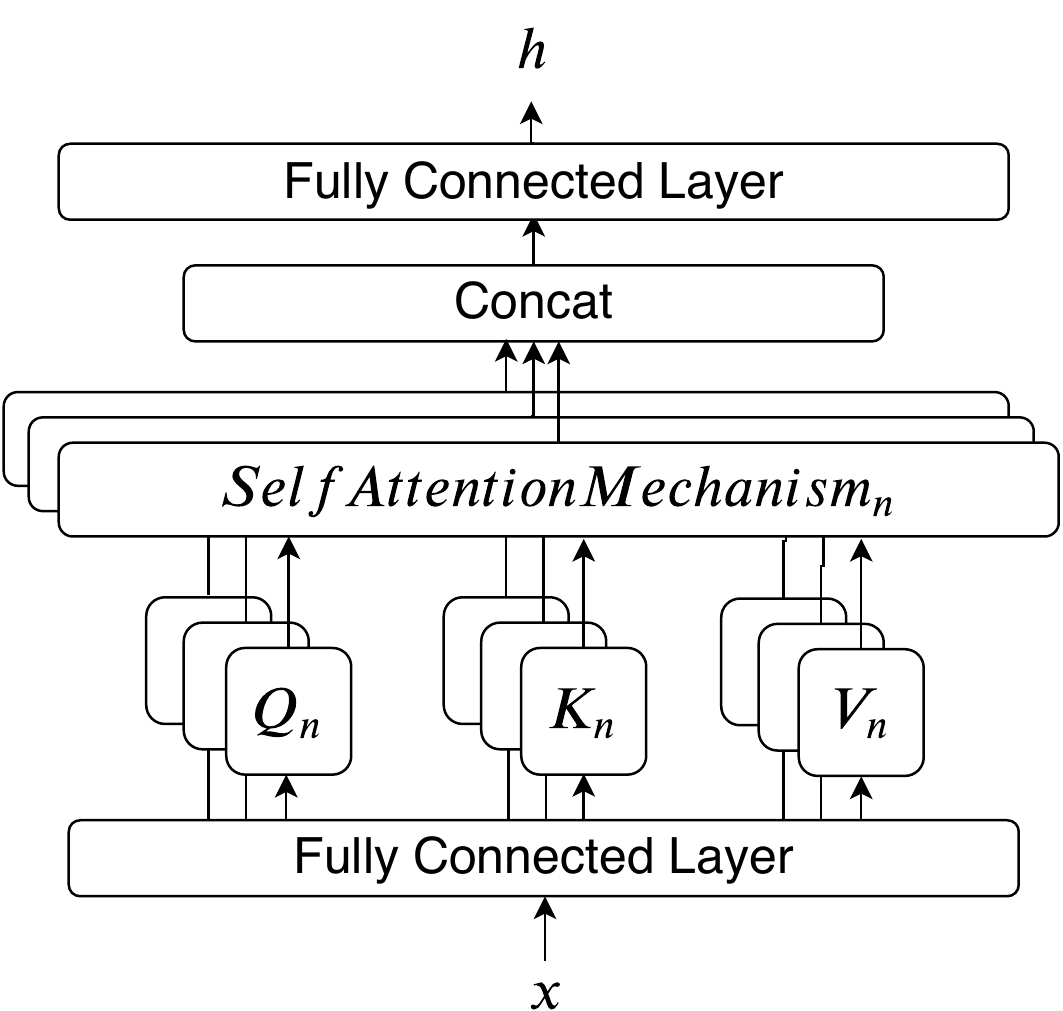} 
\caption{Diagram of attention model using Multi-Head approach using $n$ heads given an word input representation $x$ and output representation $h$.}
\label{fig:multi-attention}       
\end{figure}

To learn diverse representations of attention, we use the Multi-Head attention approach \cite{NIPS2017_7181} producing different attention scores for each word position.
We initialize each head randomly to represent different attention projections with its respective $Q$, $K$ and $V$ matrices. 
Consequently, after the training process, the Multi-Head attention approach produces a different representation subspace for each head projecting distinct attention score of each word.
To propagate its results for the subsequent layers, we concatenate all attention head results and apply a fully connected layer to reshape the output to the original size.
Figure~\ref{fig:multi-attention} illustrates how the Multi-Head approach computes each head in parallel receiving a word representation $x$ resulting in a hidden representation $h$.

\subsubsection{Position-Wise Feed Forward Network}
After computing the self-attention layer, our decoder block uses a feed-forward neural network to process each word of the sentence separately and identically. 
Instead of using a ReLU activation function as the original transformer work~\cite{NIPS2017_7181}, we follow \citeauthor{radford2018improving}~\citeyear{radford2018improving} in using the gaussian error linear unit~\cite{hendrycks2016gelu} as the activation function computed in Equation~\ref{eq:gelu}.
Equation~\ref{eq:positionwise-ff} illustrates feed-forward network (FFN) operations over each word $x$, which $W_i$ and $b_i$ represent the weight and bias of the $i$-th layer respectively.
\begin{equation} \label{eq:gelu}
	\small
    GELU(x) = 0.5x(1+tanh(\sqrt{2/\pi}(x+0.044715.x^3)
\end{equation}
\begin{equation} \label{eq:positionwise-ff}
	\small
    FFN(x) = GELU(x.W_1+b_1).W_2+b_2
\end{equation}

\subsubsection{Output Representation}
The output of a Transformer decoder is a sequence of learned embeddings of all tokens contained in the input sentence pair (i.e. the premise and hypothesis pair).
However, we need to convert the decoder output into a single representation in order to predict the class of the whole sentence pair. 
The Transformer decoder computes the embedding of this special token considering all previous words due to the masking process that we defined in attention model, which ignores the subsequent positions. 
We include a special token at the end of the input pair to represent the whole sentence since the self-attention mechanism computes its embedding considering all previous words.
With the special token embedding, we apply a fully connected layer to represent the predicted class of sentence pair.
Our model uses a softmax activation function to generate a vector of probabilities containing a single value for each class.

\section{Implementation Details}
In this section, we describe how we develop our NLI classifier.
First, we detail the NLI corpus that we use to train our neural network.
Second, we describe how we represent a premise-hypothesis pair to feed our model.
Third, we detail how we implemented our model describing the sizes of each layer and hyperparameters used.
Finally, we report how we optimize our model and measure its error reporting training results on NLI dataset.

\subsection{Stanford Natural Language Inference Corpus}
The Stanford Natural language Inference (SNLI) corpus\footnote{The corpus is free available on https://nlp.stanford.edu/projects/snli/} is a dataset that contains 570 thousand sentence pairs that were written in English and manually labeled by humans.
The pair consists of a premise and a hypothesis sentence that could follow the premise or not.
Each sentence pair contains a label that follows the NLI representation of three-way classification, which is categorized as entailment, contradiction or neutral.

Bowman \textit{et al.}~\cite{bowman:2015:snli} used Amazon Mechanical Turk for data collection by asking each worker to supply a hypothesis text based on a scene description from a pre-existing corpus~\cite{young2014image}.
After this collection step, each worker received a sentence pair and was asked to choose a single label (E for entailment, C for contradiction or N for neutral) for each pair.
This dataset contains five judgments of different workers and a consensus judgment, which is described in Table~\ref{tab:snli-example}.
Since the SNLI dataset contains descriptive phrases, all sentences are in the present tense, which is a substantial limitation for the contract conflict task, which contains multiple modal verbs.

\begin{table}[!htb]
\centering
\caption{Examples of sentence pair and judgement resulted from the data collection process~\cite{bowman:2015:snli}.}
\label{tab:snli-example}
\begin{tabular}{|l|l|c|}
\hline
\textbf{Premise Text} & \textbf{Hypothesis text} & \multicolumn{1}{l|}{\textbf{Judgements}} \\ \hline
\begin{tabular}[c]{@{}l@{}}A soccer game \\ with multiple males \\ playing a sport.\end{tabular} & \begin{tabular}[c]{@{}l@{}}Some men are \\ playing a sport.\end{tabular} & \begin{tabular}[c]{@{}c@{}}E E E E E\\ Entailment\end{tabular} \\ \hline
\begin{tabular}[c]{@{}l@{}}A black race car \\ starts up in front \\ of a crowd of \\ people.\end{tabular} & \begin{tabular}[c]{@{}l@{}}A man is driving \\ down a lonely\\  road.\end{tabular} & \begin{tabular}[c]{@{}c@{}}C C C N C\\ Contradiction\end{tabular} \\ \hline
\end{tabular}
\end{table}

\subsection{Word Representation}\label{subsubseq:input}
Given a word vocabulary, our neural network receives as input the word indexes followed by its relative position in the sentence.
Since the Transformer architecture does not have any recurrent network layer to obtain word order, we use the information of relative position to provide order meaning for our model.
We use a numeric value that represents the word position to retrieve positional embeddings as well as word embedding lookup process.

A positional embedding is a vector that contains the meaning of the position of each item in a sequence.
Instead of using fixed embeddings for each position, our model learns positional embeddings via backpropagation. 
Finally, we sum both embeddings resulting in a word representation that contains both semantic and order information feeding the following layers.

Due to our model input being composed of two sentences (premise and hypothesis), we concatenate the premise with its hypothesis followed by a special token.
This special token serves to retrieve the last hidden state, which we use as a representation of the entire pair.
Table~\ref{tab:input-example} shows an example of a premise-hypothesis contradiction pair followed by an end-of-sentence token (EOS), which the first row is the word indexes, second is the position indexes, third is the word embeddings, fourth is the position embeddings and finally the last row is the sum of embeddings.

\begin{table*}[!htb]
\centering
\caption{Example of input representation of our model composed by the meaning of word itself and its position in sentence.}
\label{tab:input-example}
\begin{tabular}{|l|l|l|l|l|l|l|l|l|l|l|l|}
\hline
Sentence pair: & A & Car & is & being & driven & . & A & Car & is & stuck & [EOS] \\ \hline
Word indexes: & 4 & 5 & 1 & 7 & 10 & 11 & 4 & 5 & 1 & 9 & 20 \\ \hline
Positional indexes: & 1 & 2 & 3 & 4 & 5 & 6 & 7 & 8 & 9 & 10 & 11 \\ \hline
Word embedding: & [1.3, ...] & [2.5, ...] & ... & ... & ... & ... & ... & ... & ... & ... & [6.7, ...] \\ \hline
Positional embedding: & [1.5, ...] & [3.7, ...] & ... & ... & ... & ... & ... & ... & ... & ... & [0.2, ...] \\ \hline
\textbf{Input representation of pair:} & \textbf{[2.8, ...]} & \textbf{[6.2, ...]} & \textbf{...} & \textbf{...} & \textbf{...} & \textbf{...} & \textbf{...} & \textbf{...} & \textbf{...} & \textbf{...} & \textbf{[6.9, ...]} \\ \hline
\end{tabular}
\end{table*}

\subsection{Neural Network Architecture Details}

We implemented our neural network model using 12 stacked blocks of Transformer decoders and included 12 heads in the attention layer of each decoder block following Radford \textit{et al}~\cite{radford2018improving}.
Considering that the SNLI dataset contains a vocabulary that holds 56220 distinct tokens, we include 56580 vectors with 240 dimensions in our embedding layer representing word embeddings including 360 vectors for position embeddings.
Although our model supports sentences with variable-length, we require to define a maximum length supported due to the embedding layer has a finite size.
We defined as maximum length supported 360 considering that the number of words in a contract clause is greater than SNLI sentences.





\subsection{Training Details}
We train our neural network with the Adam algorithm~\cite{kingma2014adam} using an initial learning rate of 6.25e-5, which decays linearly with a scheduled warmup over 0.2\% of training~\cite{loshchilov2017fixing}.
We then apply gradient clipping during optimization to avoid exploding gradients restricting the parameter's values between -1 and 1~\cite{pascanu2013difficulty}.

To measure network error, we apply the negative log-likelihood (NLL) loss function in output probabilities computed by the output layer~\cite{radford2018improving}.
Since we deal with a multi-class classification task, our loss function accumulates the log loss values of each class prediction. 
Given an output label $y_c$ for class $c$ and a premise-hypothesis pair $(p,h)$, the goal of our model is to minimize the function shown in Equation~\ref{eq:cross-entropy}.

\begin{equation}\label{eq:cross-entropy}
	\small
   NLL = -\sum_c y_{c}.\log P(c \mid p,h)
\end{equation}

\begin{figure}
\centering
  \includegraphics[width=0.5\textwidth]{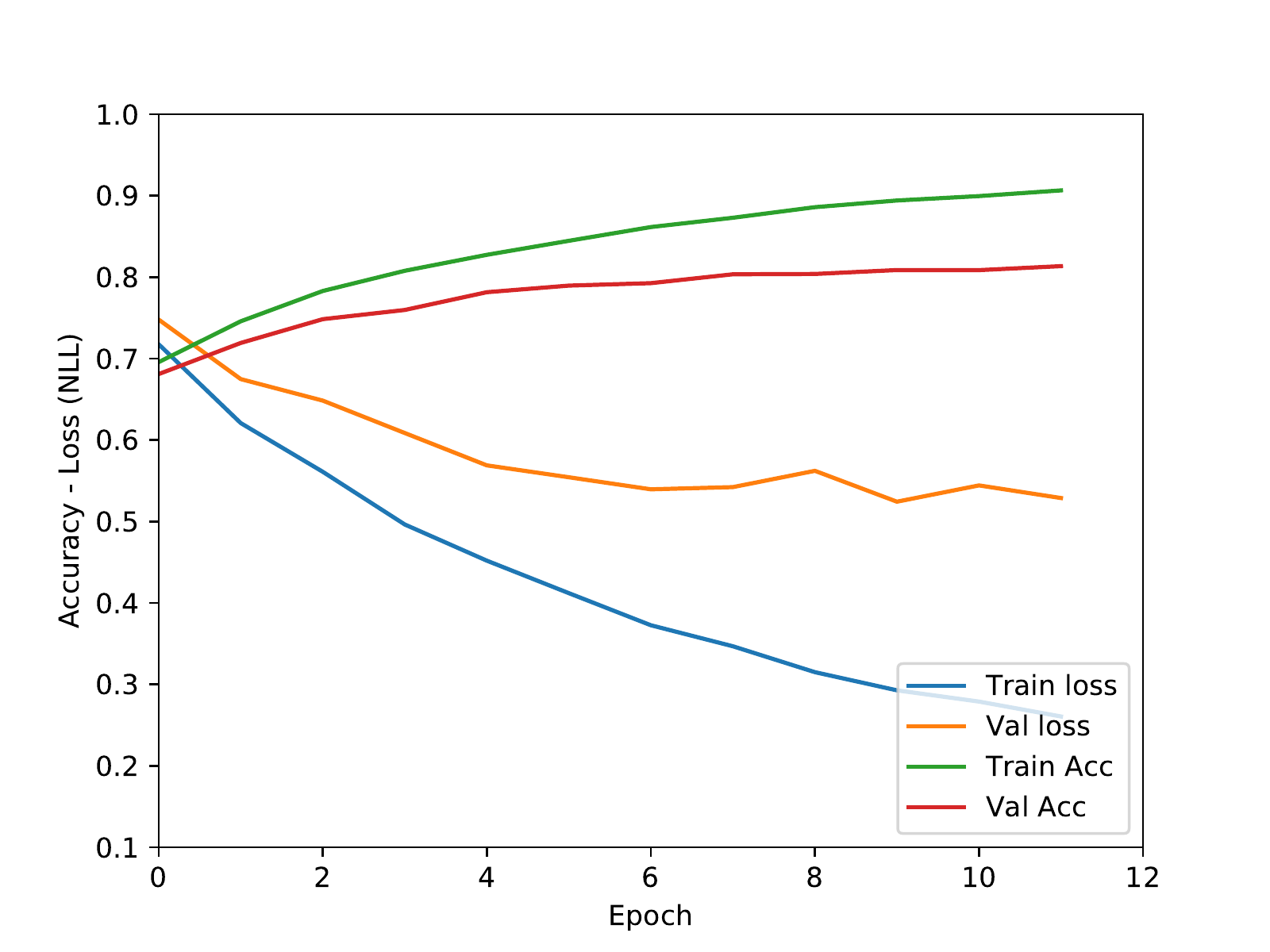}
\caption{Accuracy and loss obtained on training and validation set throughout the epochs.}
\label{fig:training-results}       
\end{figure}

During training execution, we define batches containing 16 randomly sampled instances of SNLI dataset.
We created batches with similar sizes to prevent an excess of padding through ordering samples by the sum of premise and hypothesis sentence lengths and apply early stopping to avoid overtraining our neural network using the validation set as reference.
In our training procedure, we selected as metric the accuracy obtained by our model in the validation set and 10 epochs of waiting.
Our training stopped after 22 epochs and chose the parameters with the highest accuracy obtained in epoch 12. 
We validate our training results using accuracy and loss obtained in training and validation sets throughout epochs.
Figure~\ref{fig:training-results} shows the value of the training metrics obtained by our model up to epoch 12. 

Concerning the test set, our trained neural network achieved an accuracy of 82.1\%. 
Although our work is similar to Radford \textit{et al}\cite{radford2018improving}, their work uses a pre-trained model with unsupervised learning to improve language understanding.
Table \ref{tab:result-comparison} shows details about the state-of-the-art model and our neural network concerning accuracy and number of parameters.

\begin{table}[!htb]
\centering
\caption{Results comparison of our model with Radford \textit{et al} model improved with pre-training.}
\label{tab:result-comparison}
\begin{tabular}{|l|l|l|}
\hline
 & \textbf{Radford \textit{et al}} & \textbf{Our work} \\ \hline
Training Accuracy (\%) & 96.6 & 90.5 \\ \hline
Test Accuracy (\%) & \textbf{89.9} & 82.1 \\ \hline
Number of parameters & 80m & \textbf{20m} \\ \hline
\end{tabular}
\end{table}

\section{Model Application on Conflicting Norms}

In this section, we describe the application of our trained model in conflicting norms stratified by conflict type.
First, we introduce the norm conflict dataset and explain the conflict types.
Second, we report on the results of norm pairs that contain conflicts based on conflicting modality.
Third, we report on the results of norm pairs that contain conflicts based on norm structure.
Finally, we discuss the limitations and issues of our model that we found which concern in normative actions in conflicting norm pairs.

\subsection{Norm Conflict Classification Dataset}\label{subsec:norm-dataset}

The Norm Conflict Dataset~\cite{Aires2019conflict} consists of a corpus that contains clauses from existing contracts labeled with different conflict types.
The source of these contract clauses is the Onecle\footnote{https://www.onecle.com/} site, which is a repository of business contracts.
Aires \textit{et al}~\cite{Aires2019conflict} labeled the normative sentence pairs manually using a web-based tool that selects randomly a contract clause requesting a human to create a second norm in such a way as to create conflict with the selected norm.
This dataset contains the following conflict types: \textit{deontic-modality}, \textit{deontic-structure}, \textit{deontic-object} and \textit{object-conditional}.

The \textit{deontic-modality} conflict type indicates conflicts originated by the deontic statement of each clause, i.e., prohibition $\times$ obligation, obligation $\times$ permission, and permission $\times$ prohibition.
\textit{Deontic-structure} conflict types involves different deontic meaning but with different sentence structure.
\textit{Deontic-object} conflict occurs when norm actions of the pair are conflicting, which represents the object of normative sentences.
The \textit{object-conditional} conflict occurs when condition of norm actions are conflicting.
Table \ref{tab:conflict-types} shows examples of norm pairs contained in norm conflict dataset with their respective conflict types.

\begin{table*}[!htb]
\centering
\caption{Examples of norm pairs with the respective conflict type.}
\label{tab:conflict-types}
\small
\begin{tabular}{|l|l|}
\hline
\textbf{Norm Pair} & \textbf{Conflict Type} \\ \hline
\begin{tabular}[c]{@{}l@{}}- The Specifications may be amended by the NCR design release process.\\ - The Specifications shall not be amended by the NCR design release process.\end{tabular} & deontic modality \\ \hline
\begin{tabular}[c]{@{}l@{}}- All inquiries that Seller receives on a worldwide basis relative to Buyer’s air chamber "Products" as \\ specified in Exhibit III, shall be directed to Buyer. \\ - Seller may not redirect inquiries concerning Buyer’s air chamber "Products".\end{tabular} & deontic structure \\ \hline
\begin{tabular}[c]{@{}l@{}}- Autotote shall make available to Sisal one (1) working prototype of the Terminal by May 1, 1998.\\ - Autotote shall make available to Sisal one (1) working prototype of the Terminal by June 12, 1998.\end{tabular} & deontic object \\ \hline
\begin{tabular}[c]{@{}l@{}}- The Facility shall meet all legal and administrative code standards applicable to the conduct of the \\ Principal Activity thereat.\\ - Only if previously agreed, the Facility ought to follow legal and administrative code standards.\end{tabular} & object conditional \\ \hline
\end{tabular}
\end{table*}

In this work, we apply our trained model on conflicting pairs of this dataset to find logical relations between contract clauses different conflict types.
We execute two times the model over pairs interleaving the roles of premise and hypothesis.

\subsection{Deontic Modal Conflicts}
In executions on \textit{deontic-modal} conflict pairs, we note that our model could detect the intensity of modal verbs.
Given both sentences with similar intensity, our model results in high scores of \textit{entailment} and \textit{neutrality} considering the inference direction.
For example, our model outputs the entailment relation between a considerable number of pairs that have the modal verbs \textit{shall} and \textit{may} in both directions.
However, when the premise contains the \textit{shall} verb, the entailment score is greater than premises with \textit{may}.

These scores show that our model increases entailment probability when the premise contains an obligation and the hypothesis norm contains a permissible action.
On the other hand, our model shows that the opposite is not true, which increases neutrality score when obligation comes from hypothesis norm.
However, our model fails to infer these relations when the hypothesis or premise has the modal verb \textit{will}.
Since our training dataset contains only sentences in the present tense, we suspect that our model does not recognize the word \textit{will} accurately. 
Table~\ref{tab:modal-conflict} shows examples of pairs with these modal verbs with \textit{softmax} score of our neural network output layer for neutral and entailment classes.

\begin{table*}[!htb]
\centering
\caption{Examples of norm pairs with \textit{deontic-modal} conflicts describing the model results given two norms with different modal verbs and deontic meaning.}
\label{tab:modal-conflict}
\small
\begin{tabular}{|l|l|l|l|l|}
\hline
\textbf{Norm Pair} & \multicolumn{1}{c|}{\textbf{\begin{tabular}[c]{@{}c@{}}E\\ (a,b)\end{tabular}}} & \multicolumn{1}{c|}{\textbf{\begin{tabular}[c]{@{}c@{}}N\\ (a,b)\end{tabular}}} & \multicolumn{1}{c|}{\textbf{\begin{tabular}[c]{@{}c@{}}E\\ (b,a)\end{tabular}}} & \multicolumn{1}{c|}{\textbf{\begin{tabular}[c]{@{}c@{}}N\\ (b,a)\end{tabular}}} \\ \hline
\begin{tabular}[c]{@{}l@{}}(a) Purchaser shall also be responsible for  all property taxes on the equipment.\\ (b) Purchaser may also be responsible for all property taxes on the equipment.\end{tabular} & 0.96 & 0.03 & 0.88 & 0.08 \\ \hline
\begin{tabular}[c]{@{}l@{}}(a) CoPacker shall deliver all the Products that WWI purchases under this Agreement to \\WWI F.O.B.\\ (b) CoPacker may deliver all the Products that WWI purchases under this Agreement to \\WWI F.O.B.\end{tabular} & 0.71 & 0.20 & 0.30 & 0.53 \\ \hline
\begin{tabular}[c]{@{}l@{}}(a) CBSI will retain the originals in its archives.\\ (b) CBSI may retain the originals in its archives.\end{tabular} & 0.59 & 0.35 & 0.91 & 0.07 \\ \hline
\end{tabular}
\end{table*}

In norms that contains negation between deontic meanings, our model had problems on bidirectional executions.
The trained neural network does not classify contradictions with reasonable accuracy when negations come from the premise norm.
We consider that this issue is related to our training dataset (SNLI), which may be unbalanced concerning the negation side.

\subsection{Deontic Structure}
In the \textit{deontic structure} conflict type, our model has the same problems as in \textit{deontic-modality} conflicts.
However, in this conflict case, we note that our neural network could generalize contradictions and entailment regarding modal verb in different sentence structures. 
This shows that our model can infer a logical relation where norm pairs contain different words with similar meanings. 
Table~\ref{tab:deontic-structure} shows unidirectional results of our model where the sentence (a) is the premise and sentence (b) is the hypothesis.

\begin{table*}[!htb]
\centering
\caption{Norm pairs with distinct sentence structures and their softmax scores for entailment (E), contradiction (C) and neutral (N) classes generated by our model.}
\label{tab:deontic-structure}
\small
\begin{tabular}{|l|l|l|l|}
\hline
\textbf{Norm Pair} & \multicolumn{1}{c|}{\textbf{\begin{tabular}[c]{@{}c@{}}E\\ (a,b)\end{tabular}}} & \multicolumn{1}{c|}{\textbf{\begin{tabular}[c]{@{}c@{}}C\\ (a,b)\end{tabular}}} & \multicolumn{1}{c|}{\textbf{\begin{tabular}[c]{@{}c@{}}N\\ (a,b)\end{tabular}}} \\ \hline
\begin{tabular}[c]{@{}l@{}}(a) Autotote will own the Intellectual Property Rights to all said prototypes.\\ (b) Autotote shall not own the Intellectual Property Rights to prototypes.\end{tabular} & 0.04 & 0.94 & 0.02 \\ \hline
\begin{tabular}[c]{@{}l@{}}(a) Medica shall also maintain records with respect to its costs,  obligations, and performance under \\this agreement.\\ (b) With respect to its costs, obligations, and performance under this agreement, Medica is not \\obliged to maintain records.\end{tabular} & 0.11 & 0.81 & 0.08 \\ \hline
\begin{tabular}[c]{@{}l@{}}(a) Teknika will notify LSI that it considers that a Triggering Event has occurred.\\ (b) Teknika shall not notify LSI of any regular event.\end{tabular} & 0.01 & 0.98 & 0.01 \\ \hline
\begin{tabular}[c]{@{}l@{}}(a) Customer will notify USF at least [***] days in advance of  special promotions that may cause \\unusual or excessive demand on inventory.\\ (b) Customer should notify USF of special promotions that may cause unusual and excessive \\demand on inventory.\end{tabular} & 0.91 & 0.00 & 0.09 \\ \hline
\end{tabular}
\end{table*}

\subsection{Deontic-Object and Object-Conditional}
Given conflicts that involve a difference between norm's object, our model fails to generate reasonable classifications. 
These results indicate that our neural network does not detect object context in a normative sentence on conflict context.
Based on results in both conflicts type that concern norm actions, our neural network output its prediction based on keywords such as modal verb and negation words (not). 
Therefore, our model tends to disregard norm actions when words of both norms are different or norm action structure is composed of different words.

Table~\ref{tab:object-conflicts} shows examples of norm pairs that contain conflicts related to the action of clauses.
The first example shows that our model does not recognize accurately the sentence action ignoring the associated objects.
The second example illustrates a contradiction instance that our model could capture due to the similar structure between words of pair.
Finally, the third example describes an instance that involves conflicts that concerns the conditional definition of norms.

\begin{table*}[!htb]
\centering
\caption{Example of norm pairs with conflicts that concerns in norm action with respective entailment (E), contradiction (C) and neutral (N) softmax score.}
\label{tab:object-conflicts}
\begin{tabular}{|l|l|l|l|l|}
\hline
\textbf{Norm Pair} & \multicolumn{1}{c|}{\textbf{\begin{tabular}[c]{@{}c@{}}E\\ (a,b)\end{tabular}}} & \multicolumn{1}{c|}{\textbf{\begin{tabular}[c]{@{}c@{}}C\\ (a,b)\end{tabular}}} & \multicolumn{1}{c|}{\textbf{\begin{tabular}[c]{@{}c@{}}N\\ (a,b)\end{tabular}}} & \textbf{Type} \\ \hline
\begin{tabular}[c]{@{}l@{}}(a) The arbitration shall be conducted in Tampa, Florida.\\ (b) The arbitration shall be conducted in St. Petersburg, Florida.\end{tabular} & 0.92 & 0.02 & 0.06 & deontic-object \\ \hline
\begin{tabular}[c]{@{}l@{}}(a) Hershey will cooperate in no shipping procedures.\\ (b) Hershey will cooperate in all shipping procedures.\end{tabular} & 0.05 & 0.88 & 0.07 & deontic-object \\ \hline
\begin{tabular}[c]{@{}l@{}}(a) Where applicable, Taxes shall appear as separate items on Adaptec's \\invoice.\\ (b) If shipping products, the Taxes shall appear along the other items on \\Adaptec's invoice.\end{tabular} & 0.91 & 0.04 & 0.05 & object-conditional \\ \hline
\end{tabular}
\end{table*}
\section{Related Work}
In this section, we present related work that analyze normative sentences and contract clauses.
We describe the related works explaining the problem dealt, their objectives and how they represent a normative sentence.
We compare the objective of this work with the related work and discuss the differences.


Aires \textit{et al} ~\cite{Aires2017} develop an approach that identifies potential conflicts between norms in contracts.
First, they focus on norm identification, which results in a formal representation of a norm.
Second, they use the formal representation to detect and classify potential conflicts between norms using techniques of the formal logic.
This approach assumes that a norm follows a well-defined 4-component structure: an indexing number or letter, one or more named parties, a modal verb, and a behavior description.
Given this structure, they apply a regular expression to decide whether a sentence is a norm sentence or not.
After identifying norms, they create a formal representation of the norm sentence extracting three components: party name, deontic meaning, and the norm action.
With the formal representation, they detect potential conflicts following three relations between deontic meaning in norm pairs~\cite{sadat2003methods}:
\begin{itemize}
    \item Permission and Prohibition
    \item Permission and Obligation
    \item Obligation and Prohibition
\end{itemize}
Instead of using a formal representation to use a strict logic approach, we explored the use of techniques that deal with the informal reasoning of natural language through neural network application.
We use a neural network considering SNLI dataset to deal with the challenges of NLI such as lexical semantic knowledge and the variability of linguistic expression~\cite{Maccartney:2009:NLI:1751277}.

Aires \textit{et al}~\cite{Aires2019conflict} introduce a typology of conflicts in normative sentences and present machine learning methods that classify these conflict types.
These learning methods rely on the semantic representation of norms using  Sent2Vec~\cite{pgj2017unsup} to create embedding vectors to represent the norms.
First, they describe an extension of Aires Norm Dataset to include the conflict typology, which introduces 228 new conflicting norms including the existing 111 from the previous dataset.
Second, they present an unsupervised learning method to detect the presence or absence of norm conflicts.
Finally, they present a supervised learning method that deals with binary (i.e., conflicts and non-conflict) and multi-class classification method to classify the conflict types created.
In this work, we use the dataset made by the authors to validate logical relations.
Instead of classifying conflicts, we use the conflict type to illustrating how our model can help in contract analysis showing potential conflicts that concern logical problems.
These conflicts help us to detect points that our model can improve regarding logical inference.


\section{Conclusion}
In this work, we present an approach to identify a logical relation between contract clauses.
At this point, we have trained a neural network with SNLI dataset to classify inference between a premise and a hypothesis.
We use this trained neural network on a corpus that contains a conflicting set of norms to validate whether our model can help in contract analysis.
The application of our neural network on conflicting norms could help us to identify some issues of our approach.
Although our model has issues that involve the direction of NLI inference, we show that our model can detect potential contradictions in contract clauses regardless of their structures.
Furthermore, we reported that our model can identify deontic meaning between norms assigning an entailment score based on modal verb intensity.

As future work, we intend to improve our neural network to detect the gaps identified during contract clauses analysis.
First, we intend to explore others training datasets such as the Multi-Genre NLI Corpus (MNLI)~\cite{mnli-N18-1101}, which were modeled based no SNLI but differs in that covers a range of genres of spoken and written text.
Second, we aim to use pre-trained models such as BERT~\cite{devlin2018bert}, which are state-of-the-art in a wide variety of Natural Language Process tasks such as Natural Language Inference.

\bibliographystyle{aaai}
\bibliography{bibtex}

\begin{thebibliography}{}

\bibitem[\protect\citeauthoryear{Aires \bgroup et al\mbox.\egroup
  }{2017}]{Aires2017}
Aires, J.~P.; Pinheiro, D.; Lima, V. S.~d.; and Meneguzzi, F.
\newblock 2017.
\newblock Norm conflict identification in contracts.
\newblock {\em Artificial Intelligence and Law} 25(4):397--428.

\bibitem[\protect\citeauthoryear{Aires \bgroup et al\mbox.\egroup
  }{2019}]{Aires2019conflict}
Aires, J.~P.; Granada, R.; Monteiro, J.; Barros, R.~C.; and Meneguzzi, F.
\newblock 2019.
\newblock Classification of contractual conflicts via learning of semantic
  representations.
\newblock In {\em Proceedings of the 18th International Conference on
  Autonomous Agents and Multiagent Systems (AAMAS)},  To appear.

\bibitem[\protect\citeauthoryear{Aires, Pinheiro, and
  Meneguzzi}{2017}]{aires_joao_paulo_2017_345411}
Aires, J.~P.; Pinheiro, D.; and Meneguzzi, F.
\newblock 2017.
\newblock {Norm Dataset: Dataset with Norms and Norm Conflicts}.

\bibitem[\protect\citeauthoryear{Bos and Markert}{2005}]{bos2005recognising}
Bos, J., and Markert, K.
\newblock 2005.
\newblock Recognising textual entailment with logical inference.
\newblock In {\em Proceedings of the conference on Human Language Technology
  and Empirical Methods in Natural Language Processing},  628--635.
\newblock Association for Computational Linguistics.

\bibitem[\protect\citeauthoryear{Bowman \bgroup et al\mbox.\egroup
  }{2015}]{bowman:2015:snli}
Bowman, S.; Angeli, G.; Potts, C.; and Manning, C.
\newblock 2015.
\newblock A large annotated corpus for learning natural language inference.
\newblock In {\em Conference Proceedings - EMNLP 2015: Conference on Empirical
  Methods in Natural Language Processing},  632--642.
\newblock Association for Computational Linguistics (ACL).

\bibitem[\protect\citeauthoryear{de Souza~Aires}{2015}]{AIRESDM15}
de~Souza~Aires, J.~P.
\newblock 2015.
\newblock Identifying potential conflicts between norms in contracts.
\newblock Master's thesis, Faculdade de Inform\'{a}tica -- PUCRS, Porto Alegre,
  RS, Brasil.

\bibitem[\protect\citeauthoryear{Devlin \bgroup et al\mbox.\egroup
  }{2018}]{devlin2018bert}
Devlin, J.; Chang, M.-W.; Lee, K.; and Toutanova, K.
\newblock 2018.
\newblock Bert: Pre-training of deep bidirectional transformers for language
  understanding.
\newblock {\em arXiv preprint arXiv:1810.04805}.

\bibitem[\protect\citeauthoryear{Hendrycks and
  Gimpel}{2016}]{hendrycks2016gelu}
Hendrycks, D., and Gimpel, K.
\newblock 2016.
\newblock Gaussian error linear units (gelus).
\newblock {\em arXiv preprint arXiv:1606.08415}.

\bibitem[\protect\citeauthoryear{Kingma and Ba}{2014}]{kingma2014adam}
Kingma, D.~P., and Ba, J.
\newblock 2014.
\newblock Adam: A method for stochastic optimization.
\newblock {\em arXiv preprint arXiv:1412.6980}.

\bibitem[\protect\citeauthoryear{Loshchilov and
  Hutter}{2017}]{loshchilov2017fixing}
Loshchilov, I., and Hutter, F.
\newblock 2017.
\newblock Fixing weight decay regularization in adam.
\newblock {\em arXiv preprint arXiv:1711.05101}.

\bibitem[\protect\citeauthoryear{Maccartney}{2009}]{Maccartney:2009:NLI:1751277}
Maccartney, B.
\newblock 2009.
\newblock {\em Natural Language Inference}.
\newblock Ph.D. Dissertation, Stanford, CA, USA.
\newblock AAI3364139.

\bibitem[\protect\citeauthoryear{Pagliardini, Gupta, and
  Jaggi}{2018}]{pgj2017unsup}
Pagliardini, M.; Gupta, P.; and Jaggi, M.
\newblock 2018.
\newblock {Unsupervised Learning of Sentence Embeddings using Compositional
  n-Gram Features}.
\newblock In {\em NAACL 2018 - Conference of the North American Chapter of the
  Association for Computational Linguistics}.

\bibitem[\protect\citeauthoryear{Pascanu, Mikolov, and
  Bengio}{2013}]{pascanu2013difficulty}
Pascanu, R.; Mikolov, T.; and Bengio, Y.
\newblock 2013.
\newblock On the difficulty of training recurrent neural networks.
\newblock In {\em International conference on machine learning},  1310--1318.

\bibitem[\protect\citeauthoryear{Radford \bgroup et al\mbox.\egroup
  }{2018}]{radford2018improving}
Radford, A.; Narasimhan, K.; Salimans, T.; and Sutskever, I.
\newblock 2018.
\newblock Improving language understanding by generative pre-training.
\newblock {\em URL https://s3-us-west-2. amazonaws.
  com/openai-assets/research-covers/languageunsupervised/language understanding
  paper. pdf}.

\bibitem[\protect\citeauthoryear{Sadat-Akhavi}{2003}]{sadat2003methods}
Sadat-Akhavi, A.
\newblock 2003.
\newblock {\em Methods of resolving conflicts between treaties}, volume~3.
\newblock Martinus Nijhoff Publishers.

\bibitem[\protect\citeauthoryear{Sutskever, Vinyals, and
  Le}{2014}]{sutskever2014sequence}
Sutskever, I.; Vinyals, O.; and Le, Q.~V.
\newblock 2014.
\newblock Sequence to sequence learning with neural networks.
\newblock In {\em Advances in neural information processing systems},
  3104--3112.

\bibitem[\protect\citeauthoryear{Vaswani \bgroup et al\mbox.\egroup
  }{2017}]{NIPS2017_7181}
Vaswani, A.; Shazeer, N.; Parmar, N.; Uszkoreit, J.; Jones, L.; Gomez, A.~N.;
  Kaiser, L.~u.; and Polosukhin, I.
\newblock 2017.
\newblock Attention is all you need.
\newblock In Guyon, I.; Luxburg, U.~V.; Bengio, S.; Wallach, H.; Fergus, R.;
  Vishwanathan, S.; and Garnett, R., eds., {\em Advances in Neural Information
  Processing Systems 30}. Curran Associates, Inc.
\newblock  5998--6008.

\bibitem[\protect\citeauthoryear{Williams, Nangia, and
  Bowman}{2018}]{mnli-N18-1101}
Williams, A.; Nangia, N.; and Bowman, S.
\newblock 2018.
\newblock A broad-coverage challenge corpus for sentence understanding through
  inference.
\newblock In {\em Proceedings of the 2018 Conference of the North American
  Chapter of the Association for Computational Linguistics: Human Language
  Technologies, Volume 1 (Long Papers)},  1112--1122.
\newblock Association for Computational Linguistics.

\bibitem[\protect\citeauthoryear{Young \bgroup et al\mbox.\egroup
  }{2014}]{young2014image}
Young, P.; Lai, A.; Hodosh, M.; and Hockenmaier, J.
\newblock 2014.
\newblock From image descriptions to visual denotations: New similarity metrics
  for semantic inference over event descriptions.
\newblock {\em Transactions of the Association for Computational Linguistics}
  2:67--78.

\end{thebibliography}

\end{document}